\documentclass[conference]{IEEEtran}

\usepackage{xspace}
\usepackage{graphicx}
\usepackage{amsmath,amssymb}
\usepackage{natbib} 
\usepackage{booktabs}
\usepackage{hyperref}

\newcommand{\topclustrag}[0]{\textsc{TopClustRAG}\xspace}
\begin{document}

\title{TopClustRAG at SIGIR 2025 LiveRAG Challenge}

\author{
    \IEEEauthorblockN{Juli Bakagianni\IEEEauthorrefmark{1}, John Pavlopoulos\IEEEauthorrefmark{1}\IEEEauthorrefmark{2}, Aristidis Likas\IEEEauthorrefmark{3}}
    \IEEEauthorblockA{\IEEEauthorrefmark{1}Athens University of Economics and Business, Greece\\
    Email: \{julibak, ipavlopoulos\}@aueb.gr}
    \IEEEauthorblockA{\IEEEauthorrefmark{2}Archimedes, Athena Research Center, Greece}
    \IEEEauthorblockA{\IEEEauthorrefmark{3}Computer Science and Engineering, University of Ioannina, Greece\\
    Email: arly@cs.uoi.gr}
}





\maketitle
\begin{abstract}
We present \topclustrag, a retrieval-augmented generation (RAG) system developed for the LiveRAG Challenge, which evaluates end-to-end question answering over large-scale web corpora. Our system employs a hybrid retrieval strategy combining sparse and dense indices, followed by K-Means clustering to group semantically similar passages. Representative passages from each cluster are used to construct cluster-specific prompts for a large language model (LLM), generating intermediate answers that are filtered, reranked, and finally synthesized into a single, comprehensive response. This multi-stage pipeline enhances answer diversity, relevance, and faithfulness to retrieved evidence. Evaluated on the FineWeb Sample-10BT dataset, \topclustrag ranked 2nd in faithfulness and 7th in correctness on the official leaderboard, demonstrating the effectiveness of clustering-based context filtering and prompt aggregation in large-scale RAG systems.
\end{abstract}

\maketitle

\section{Introduction}

Retrieval-Augmented Generation (RAG) has emerged as a promising paradigm to enhance the factuality and contextual grounding of large language models (LLMs), particularly in open-domain question answering. By supplementing generation with document retrieval, RAG systems aim to produce responses that are both informative and faithful to source material. However, designing effective RAG pipelines remains challenging due to issues such as noisy retrieval results, redundant or semantically overlapping passages, and the difficulty of aggregating diverse evidence into a coherent response.

To advance research in this domain, the Technology Innovation Institute (TII) organized the LiveRAG Challenge,\footnote{\url{https://liverag.tii.ae/}} a leaderboard-based competition that evaluates end-to-end RAG systems at scale. Participants were tasked with building retrieval-augmented QA systems over the FineWeb Sample-10BT corpus~\cite{penedo2024the}, a 10-billion-token web dataset, and were evaluated on correctness and faithfulness using metrics derived from LLM-based automatic assessments. Systems were required to operate in a low-latency setting, with strong emphasis on faithfulness to retrieved evidence and informativeness of generated answers.

In this paper, we present \topclustrag, our submission to the LiveRAG Challenge. \topclustrag is a multi-stage RAG system designed to improve answer quality through hybrid retrieval, clustering-based content selection, and prompt-based synthesis. Our pipeline clusters semantically similar passages retrieved via a hybrid sparse-dense index, selects representative passages from each cluster, and prompts an LLM to generate intermediate answers. These responses are then filtered, re-ranked, and synthesized into a final answer. Through this design, \topclustrag balances retrieval diversity with answer precision.

Our system ranked \textbf{2nd in faithfulness} and \textbf{7th in correctness} in the first session of the leaderboard, demonstrating that cluster-based prompt diversification and re-ranking can yield highly grounded and accurate responses. We provide a detailed description of our system components, synthetic evaluation setup, and leaderboard performance in the following sections.

\section{System Description}
In the following sections, we present an overview of our system, \topclustrag, along with the key components of its pipeline.

\subsection{System Overview}

\begin{figure*}
    \centering
    \includegraphics[trim={0 0 0 110},clip,width=\textwidth]{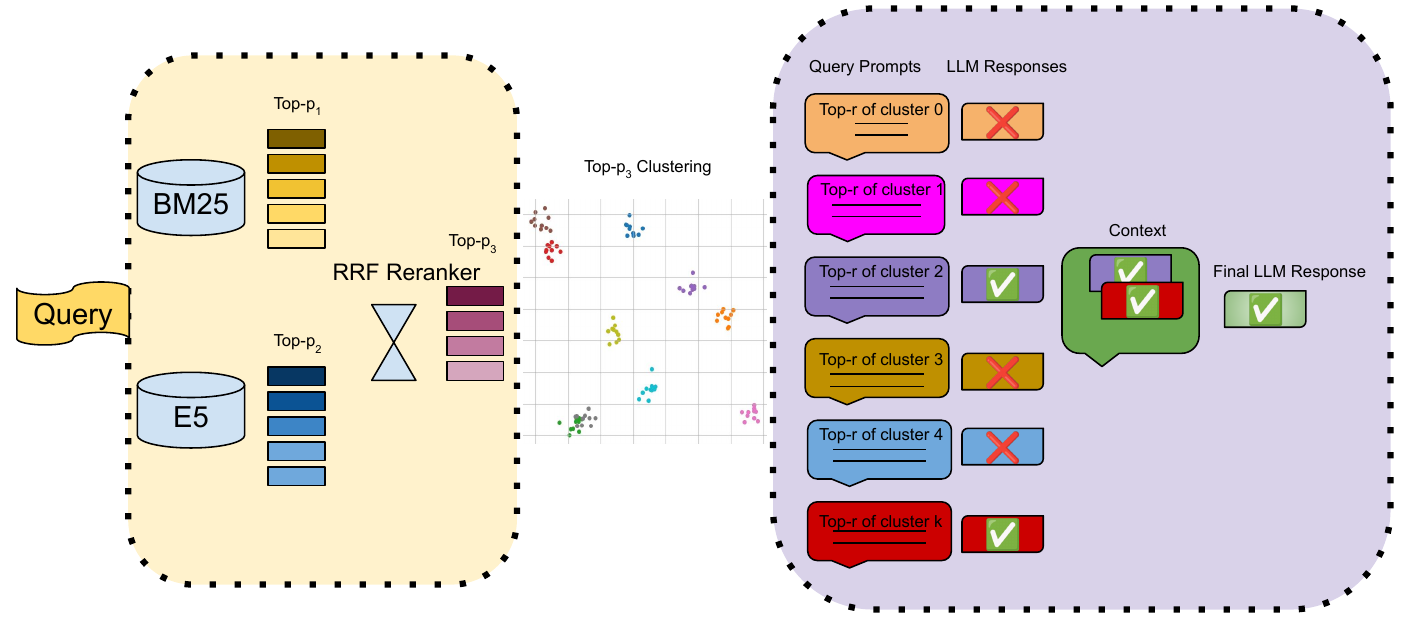}
    \caption{Overview of the \topclustrag pipeline.}
    \label{fig:system}
\end{figure*}

Figure~\ref{fig:system} illustrates the architecture of our proposed RAG system, \topclustrag. The system employs a multi-stage pipeline designed to enhance answer quality through hybrid retrieval, clustering-based context filtering, and prompt augmentation.

The process begins by retrieving the top \( p \) passages for a given query using a hybrid retriever that combines results from both sparse (BM25) and dense (embedding-based) indices. The final retrieval scores are computed using Reciprocal Rank Fusion (RRF), and the top \( p \) ranked passages form the candidate pool.

To reduce redundancy and emphasize relevant content, we apply K-Means clustering to the TF-IDF representations of the top-\( p \) passages. The optimal number of clusters \( k \) is selected by maximizing the average macro-silhouette coefficient \cite{pavlopoulos2024revisiting}, which is robust to class imbalance.

From each cluster, we select the top \( r \) representative passages based on their original hybrid retrieval scores, reflecting their relevance to the query. Each set of representative passages from a cluster is used as context in a prompt to Falcon3-10B-instruct LLM \cite{Falcon3}, alongside the original query. The LLM is asked to generate an answer if the required information is present in the given context.

The generated candidate answers (one per cluster) are filtered to retain only those that contain substantive responses.\footnote{We exclude the ``I don't know.'' responses.} These are then reranked using the cross-encoder/ms-marco-MiniLM-L6-v2\footnote{\url{https://huggingface.co/cross-encoder/ms-marco-MiniLM-L6-v2}} CrossEncoder reranker. Finally, all reranked answers are provided to the LLM in an aggregated prompt, instructing it to synthesize a single, final answer based on the combined evidence.

\subsection{Synthetic Data Generation}

The LiveRAG Challenge is based on the FineWeb Sample-10BT dataset \cite{penedo2024the}, a collection of 14.9 million documents and 10 billion tokens randomly sampled from FineWeb. This dataset includes diverse web content, such as news articles, blogs, academic texts, and product descriptions.

As the challenge organizers did not provide a validation set, we constructed a synthetic validation dataset using the method proposed by \citet{filice2025generating}, applied to the FineWeb Sample-10BT collection. The resulting dataset comprises 27 entries, covering 22 unique question categories and six distinct user categories. Notably, three of the question–answer pairs are supported by two different documents from the dataset, allowing for multi-source reasoning.

\subsection{System Modules}
\subsubsection{Retrieval System}\label{sec:retrieval}
Our retrieval system builds upon the pre-constructed indices provided by the challenge organizers: a sparse index based on BM25 (via OpenSearch)\footnote{\url{https://opensearch.org/}} and a dense index based on intfloat/e5-base-v2 \cite{wang2022text} hosted on Pinecone.\footnote{\url{https://www.pinecone.io/}} In addition to evaluating each retriever individually, we explored a hybrid retrieval strategy that combines the outputs of both using Reciprocal Rank Fusion (RRF). For each query, documents received fused scores based on their reciprocal ranks in the sparse and dense results, and were subsequently re-ranked to produce a unified top-\(k\) list.

To select the most suitable retrieval method, we evaluated the three systems on the synthetic validation dataset using standard information retrieval metrics. Specifically, we report Mean Reciprocal Rank (MRR)~\cite{10.5555/553876}, which captures the average inverse rank of the first relevant document, and Recall at rank \(k\) (R@1, R@5, R@10, R@50, R@100, R@200, R@1000), which measures whether at least one relevant document is retrieved within the top \(k\) results.

As shown in Table~\ref{tab:ir_metrics}, the sparse retriever outperforms both the dense and hybrid methods in terms of MRR and R@1 through R@10, indicating superior early precision. However, the hybrid retriever achieves the highest recall at deeper ranks (R@50 and beyond), which is critical for our system, as it benefits from access to relevant passages beyond the top-10. Based on this trade-off, we selected the hybrid retrieval approach for the final system deployment.

\begin{table}
  \caption{Retrieval performance of sparse, dense, and hybrid systems on the synthetic validation set.}
  \label{tab:ir_metrics}
  \resizebox*{\columnwidth}{!}{
\begin{tabular}{lcccccccc}
\toprule
System  & MRR & R@1 & R@5 & R@10 & R@50 & R@100 & R@200 & R@1000 \\
\midrule
Sparse  & \textbf{0.3361} & \textbf{0.2037} & \textbf{0.4074} & \textbf{0.4815} & 0.5741 & 0.6481 & 0.7593 & 0.8704 \\
Dense   & 0.0526 & 0.0000 & 0.0926 & 0.1111 & 0.2963 & 0.3333 & 0.3519 & 0.5556 \\
Hybrid  & 0.1322 & 0.0370 & 0.1111 & 0.3519 & \textbf{0.6852} & \textbf{0.7778} & \textbf{0.8519} & \textbf{0.8889} \\
\bottomrule
\end{tabular}
}
\end{table}

\subsubsection{Passage Embedding and Dimensionality Reduction}\label{sec:emb}

To enable clustering over the retrieved passages, we first represent each passage using TF-IDF vectors, which capture the importance of terms across the corpus. Given the high dimensionality of TF-IDF representations, we apply Singular Value Decomposition (SVD) to reduce each vector to 100 dimensions. This dimensionality reduction step helps preserve semantic structure while improving computational efficiency.

We opted for TF-IDF embeddings primarily due to their speed, which was a key consideration in the challenge setting. For comparison, we also experimented with contextual embeddings derived from the billatsectorflow/stella\_en\_1.5B\_v5 model~\citep{zhang2025stella}, a language model that offers a favorable trade-off in terms of MTEB performance,\footnote{\url{https://huggingface.co/spaces/mteb/leaderboard}} memory consumption, and token capacity.

Table~\ref{tab:emb-results} shows that \topclustrag achieves comparable—and in terms of BERTScore F1, even superior—performance using the simpler TF-IDF representations. This empirical result justifies our choice of TF-IDF as the default embedding method, balancing efficiency and effectiveness.

\begin{table}[h]
\centering
\caption{ROUGE-L and BERTScore F1 of \topclustrag\ using contextual (Stella 1.5b) vs.\ TF-IDF representations.}
\label{tab:emb-results}
\resizebox{\columnwidth}{!}{
\begin{tabular}{lcc}
\toprule
\textbf{Embedding Type} & \textbf{ROUGE-L} & \textbf{BERTScore F1} \\
\midrule
contextual embeddings & \bf 0.217 & 0.531\\
TF-IDF & 0.214 & \bf 0.546\\
\bottomrule
\end{tabular}
}
\end{table}

\subsubsection{Clustering of Retrieved Passages}\label{sec:clustering}

To group semantically similar passages, we apply K-Means clustering to the dimensionality-reduced passage embeddings. The number of clusters \(k\) is not fixed a priori; instead, it is selected dynamically for each query by maximizing the macro-averaged silhouette coefficient, a metric that captures both intra-cluster cohesion and inter-cluster separation. This approach ensures robustness to potential class imbalance in the distribution of relevant content across clusters~\cite{pavlopoulos2024revisiting}.
This clustering step serves two purposes: it filters out redundant content, and it enables the generation of diverse candidate answers by isolating different semantic aspects of the retrieved content.


\subsubsection{Selection of Retrieved Passages}\label{sec:select-passages}

We select \( r \) representative passages per cluster to serve as context for answer generation. We compare two strategies for this selection: (i) ranking passages by their original hybrid retrieval scores (reflecting query relevance), and (ii) ranking by proximity to the cluster centroid. 
Table~\ref{tab:combined_metrics} shows that selecting passages based on retrieval scores significantly outperforms centroid-based selection in Recall. Consequently, we adopt score-based selection as the default strategy. 
We determine the optimal value of \( r \) to be 5, because  adding more than 5 passages (R@5-R@10) does not improve Recall.

\begin{table}[ht]
\centering
\caption{Recall@r for passage selection strategies, by query relevance \textbf{score} or by \textbf{proximity} to the cluster centroid (for clusters containing at least one gold passage).}
\label{tab:combined_metrics}
\resizebox{\columnwidth}{!}{
\begin{tabular}{lcccccc}
\toprule
Metric & R@1 & R@2 & R@3 & R@4 & R@5 & R@10 \\
\midrule
Distance & 0.00 & 0.04 & 0.19 & 0.26 & 0.30 & 0.48 \\
Score    & \textbf{0.48} & \textbf{0.67} & \textbf{0.70} & \textbf{0.85} & \textbf{0.93} & \textbf{0.93} \\
\bottomrule
\end{tabular}
}
\end{table}

\subsubsection{Cluster-Based Prompt Construction}\label{sec:prompt-constr}

For each cluster, the selected representative passages are concatenated to form a single prompt. This results in one prompt per cluster, each tailored to a specific semantic grouping of the retrieved content. This approach preserves topical diversity and enhances relevance by isolating distinct thematic components. Table~\ref{tab:instruction} shows the instruction template used for prompt construction.

\begin{table}[ht]
    \centering
    \caption{Instruction template for RAG.}
    \label{tab:instruction}
    \resizebox{\columnwidth}{!}{
    \setlength{\fboxrule}{0.5pt} 
    \fbox{%
        \begin{tabular}{p{\linewidth}}
            Answer the question using only the context below. Do not make up any new information.
            If no part of the answer is found in the context, respond only with: \texttt{``I don't know.''}
            If only part of the answer is found, include that part in a complete sentence that uses the phrasing of the question, and state that the rest is not available in the context.
            If the full answer is found, respond with a complete sentence that includes the phrasing of the question.

            Context:

            \texttt{<1st text>}

            \texttt{<2nd text>}

            \texttt{<5th text>}

            Question: \texttt{<question>}

            Answer:
        \end{tabular}
    }%
    }
\end{table}

\subsubsection{Intermediate Response Generation}\label{sec:resp-gen}

Each cluster-specific prompt is independently processed by the Falcon3-10B-instruct LLM to generate one intermediate response per cluster. This step enables the model to generate focused answers grounded in diverse subsets of the retrieved passages, increasing coverage of relevant content across semantic clusters.

\subsubsection{Final Response Synthesis}\label{sec:final-resp}

Intermediate responses are first filtered to remove non-informative outputs, meaning those containing ``I don't know.'' The remaining responses are then scored using the \texttt{cross-encoder/ms-marco-MiniLM-L6-v2} model, which evaluates their relevance to the original query. The top-ranked responses are concatenated into a final aggregated context, which is used as input for a final prompt to the language model. The instruction template employed is the same as in the cluster-based prompt construction step, enabling the model to generate a single, comprehensive answer based on consolidated evidence.

\section{Experimental Setup}

All experiments, including the Live Challenge Day session, were conducted on Google Colab using CPU resources. The language model employed was Falcon3-10B-Instruct running on the AI71 platform,\footnote{\url{https://ai71.ai/}} as provided by the challenge organizers. Retrieval leveraged the prebuilt indices from OpenSearch (sparse) and Pinecone (dense). For each question, we retrieved 200 passages from OpenSearch and 200 from Pinecone. These results were combined using our hybrid retrieval strategy, and the top 100 passages were retained for downstream processing.

The LiveRAG Challenge Day event lasted two hours and consisted of 500 questions. To maximize throughput, we utilized ten parallel requests per query and eight parallel AI71 API clients. Our system processed all 500 questions within approximately one hour.

\section{Evaluation on Leaderboard}

We submitted our system to the leaderboard evaluation, which ranks participating systems based on two metrics: \textbf{Correctness} and \textbf{Faithfulness}. 
The former ranges from -1 to 2 and combines two components: (i) \emph{coverage}, defined as the proportion of vital information—identified by a strong LLM—in the ground truth answer that is present in the generated response~\cite{pradeep2025great}; and (ii) \emph{relevance}, which measures the extent to which the generated response directly addresses the question, regardless of factual correctness.
The latter ranges from -1 to 1 and evaluates whether the generated response is grounded in the retrieved passages~\cite{es2024ragas}.
We participated in the first evaluation round, ranking \textbf{7th in correctness} with a score of 0.685146, and \textbf{2nd in faithfulness} with a score of 0.460062.
Out of 500 evaluation questions, our system responded with ``I don't know.'' to 109 questions—receiving a zero score for both metrics on those instances—while it generated substantive answers for the remaining queries.

\section{Experiments}

\paragraph{Baselines} We compare our \topclustrag system three baselines. \textbf{The first baseline} uses the top-$k$ passages ranked by the retrieval score, where $k \in \{0,1,5\}$, representing common values. \textbf{The second baseline} is an ablation of our system that ignores clustering. It retrieves ten passages according to the retrieval score, which are split into five batches of five passages, then processed following the \topclustrag steps~\ref{sec:prompt-constr}--\ref{sec:final-resp}--without clustering.
\textbf{The third baseline} extends the second baseline by selecting an optimal $k$ value per query, following the approach \topclustrag. For each query, $k$ is defined by the number of clusters (see the clustering step of \S\ref{sec:clustering}) multiplied by five (see \S\ref{sec:select-passages}).

\paragraph{Evaluation metrics} Evaluation is performed using ROUGE-L and BERTScore F1 and Table~\ref{tab:results} summarizes the results for the baselines against our \topclustrag system. We observe that when no passages are retrieved (top-0), ROUGE-L is near zero while BERTScore F1 is relatively higher. In this setting, 25 out of 27 responses return ``I don't know'', which means that BERTScore may capture general fluency or semantic plausibility even in the absence of content-specific grounding. However, even though we observe that both ROUGE-L and BERTScore are not ideal for this task---since answers may be found across multiple documents and not solely in the gold reference(s)---they offer useful insights when comparing systems under the same input constraints.

\paragraph{Results} The top three rows of Table~\ref{tab:results} show that \topclustrag outperforms the first baseline for common $k$ values. The second baseline (top-10), which uses batching and employs an LLM, \topclustrag is overall better, performing slightly worse in ROUGE-L but better in BERTScore F1. The third baseline, i.e., the top-$5k$ ablation, is better than top-5 (of the 1st baseline) but worse than top-10 (2nd baseline) and \topclustrag. Although it uses approximately the same number of passages as \topclustrag (without clustering), 
performance is worse likely due to \textbf{diminishing returns} (additional passages contribute noise) and \textbf{lack of structure} (passages are not organized to ensure diverse topical coverage). This suggests that clustering contributes meaningfully beyond simple context size expansion. Specifically, the structured \textbf{selection of representative passages} from each cluster (rather than random or purely relevance-based sampling) helps ground the generation process in more diverse and complementary information.
We note that there may be $k$ values for which the 1st baseline could be the best, but the selection of the best $k$ value is arbitrary and cannot guarantee robustness.

\begin{table}[h]
\centering
\caption{ROUGE-L and BERTScore F1 for the three baselines, top-$k$ with $k \in \{0,1,5\}$, top-10, and top-$5k$, compared to our \topclustrag. The best results are shown in bold.}
\label{tab:results}
\resizebox{\columnwidth}{!}{
\begin{tabular}{lcc}
\toprule
\textbf{} & \textbf{ROUGE-L} & \textbf{BERTScore F1} \\
\midrule
top-0 &  0.031 & 0.317 \\
top-1 &  0.186 &0.494 \\
top-5  & 0.202 & 0.515\\
\hline
top-10  & \bf 0.217 & 0.541\\\hline
top-$5k$ & 0.202 &	0.532\\\hline
TopClustRAG & 0.214 &\bf 0.546\\
\bottomrule
\end{tabular}
}
\end{table}

\section{Conclusions}

In this study, we presented \topclustrag, our RAG system designed for large-scale question answering over diverse web data in the LiveRAG Challenge. By leveraging a hybrid sparse-dense retrieval approach combined with clustering-based filtering, our system effectively balances retrieval diversity and relevance. The proposed multi-stage pipeline—comprising cluster-based prompt construction, intermediate answer generation, re-ranking, and final synthesis—enhances the faithfulness and informativeness of the generated responses.

Our evaluation on the FineWeb Sample-10BT dataset demonstrated that \topclustrag achieves competitive leaderboard performance, ranking 2nd in faithfulness and 7th in correctness in the first session. These results highlight the potential of clustering and prompt aggregation techniques to improve large language model grounding in retrieval-augmented generation. In the future, we plan to extend clustering to handle larger numbers of retrieved passages, assessing whether clustering can help uncover relevant information that may be hidden within the extensive data.

\bibliographystyle{ACM-Reference-Format}
\bibliography{sample-base,custom}


\end{document}